%% file: compact.tex
\documentclass[11pt]{article}
\usepackage[round]{natbib}
\usepackage{times}

\input{defn.tex}

  \usepackage{graphicx}

\renewcommand{\cite}{\citep}
\renewcommand{\citeyear}{\citeyearpar}
\renewcommand{\S}{{\cal S}}

\newcommand{\ML}{\mathit{M}}

\renewcommand{\FF}{\mathit{FF}}

\newcommand{\AP}{\mathit{AP}}
\newcommand{\BP}{\mathit{BP}}
\newcommand{\AO}{\mathit{AO}}
\newcommand{\PD}{\mathit{PO}}

\newcommand{\WD}{\mathit{WD}}

\renewcommand{\gets}{=}
\newcommand{\Pl}{\mbox{Pl}}
\newcommand{\Dom}{\mbox{\it Dom}}

\begin{document}
\title{Compact Representations of 
Extended
Causal Models}

\author{
Joseph Y. Halpern\thanks{Supported in part by NSF grants 
IIS-0812045 and IIS-0911036, AFOSR grants FA9550-08-1-0438 and
AFOSR grants FA9550-08-1-0438 and
FA9550-05-1-0055, and ARO grant W911NF-09-1-0281.} 
\\
Cornell University\\
halpern@cs.cornell.edu
\and
Christopher Hitchcock\\
California Institute of Technology\\
Cricky@caltech.edu}
\date{October 2012}
\maketitle

\begin{abstract}
Judea Pearl \citeyear{pearl:2k} was the first to propose a definition of \emph{actual causation}
using causal models. A number of authors have suggested that an adequate 
account of actual causation must appeal not only to causal structure, but also
to considerations of \emph{normality}. In \cite{HH11}, we offer a
definition of actual causation  
using \emph{extended causal models}, which include information about both 
causal structure and normality. Extended causal models are potentially very complex.
In this paper, we show how it is possible to achieve a compact representation of
extended causal models.
\end{abstract} 

\section{Introduction}

One of Judea Pearl's many, many important contributions 
to the study of causality
was the first attempt to use the mathematical tools of causal modeling
to give an account of ``actual causation'', a notion that has been of
considerable interest among philosophers and legal theorists  
\cite[Chapter 10]{pearl:2k}.  Pearl later revised his account of actual
causation in joint work with Halpern \cite{HP01b}.   A number of
authors \cite{Hall07,Hal39,Hitchcock07,Menzies04}  
have suggested that an account of actual causation must be sensitive to considerations of normality, as well as to causal structure. In 
\cite{HH11},
we suggest a way of incorporating considerations of normality into the Halpern-Pearl theory, and show how to extend the account to illuminate features of the psychology of causal judgment, as well as features of 
causal
reasoning in the law. 
Our account of actual causation 
makes use of ``extended causal models'', which include
both structural equations
among a set of variables,
and a partial preorder on possible
worlds, which 
represents the relative
``normality'' 
of those worlds.

We actually want to think of people as working with the structural
equations and normality order to evaluate 
actual causation.
However, 
consideration of even simple examples immediately suggests a problem. 
A direct representation of the equations and normality order
is too cumbersome for cognitively limited agents to use effectively.
If our account of actual causation is to be at all realistic as a model of
human causal judgment, some form of compact representation will be needed.


To understand the problem, consider a doctor trying to deal with a
patient who has just come in reporting bad headaches. Let's keep things
simple, and suppose that the doctor considers only a small number
of variables that might be relevant: stress, constriction of blood
vessels in the brain, aspirin consumption, and trauma to the
head. Again, keeping things simple, assume that each of these variables
(including headaches) is binary; that is, has only two possible values.
So, for example, the patient either has a headache or not. 
Each variable may depend upon the value of the other four.
To represent that structural equation for the variable ``headaches'', a
causal model will need to assign a value to ``headache'' for each of the
sixteen possible  
values of the other four variables. That means that there are 
$2^{16}$---over 60,000!---%
possible equations for ``headaches''. Considering all five variables,
there are $2^{80}$ (over $10^{24}$) possible sets of
equations. Representing one of these  
would require eighty binary bits of information.
Now consider the normality orders. 
With five
binary variables, there are $2^5=32$ possible assignments of values to
these variables. Think of each of these assignments as a ``possible
world''. There are $32!$ (roughly $2.6 \times 10^{35}$) strict orders
of these 
32 worlds, and many more if we allow for ties or incomparable worlds. 
Altogether, the doctor would need to store close to two hundred 
bits of information just to represent this simple extended causal model.

Now suppose we consider a more realistic model with 50 random variables.
Then the same arguments 
show that we would need as 
many as
$2^{50 \times 2^{49}}$
possible sets of equations, $2^{50}$ possible worlds, and over
$2^{50 \times 2^{50}}$ normality orders (in general, with $n$ binary
variables, 
there are 
$2^{n2^{n-1}}$
sets of equations, $2^n$ possible worlds, and
$(2^n)! \sim 2^{n2^n}$ strict orders).  Thus, with 50 variables, roughly
$50 \times 2^{50}$ bits would be needed to representing a causal model.
This is clearly cognitively unrealistic. 

The goal of this paper is to show that, in practice, representing 
the information needed to evaluate 
actual causation
can be done in a
reasonably compact way, so that the assumption that people are actually
doing this is indeed psychologically plausible. To do this, we will
make significant 
use of another of Pearl's signal contributions: the use of directed
graphs---specifically, \emph{Bayesian networks}---to represent
independence relations  
\cite{Pearl}.

The first step towards the goal of getting a compact representation comes from the observation that similar representational difficulties arise when it comes to reasoning 
about probability.  For example, if the doctor would like to reason
probabilistically about the symptoms, just describing a probability
distribution on the 
$2^{50}$ 
worlds would also require 
$2^{50}$
(or,
more precisely, $2^{50} - 1$) numbers.   Bayesian networks allow us to
typically get 
much more compact representations of probability distributions by taking
advantage of (conditional) independencies.  

Our goal is to arrive at an analogue of a Bayesian network
representation for both the structural equations and for normality.   
For the structural equations, it is easy to see where independence comes
in.  If the equation for each variable $X$ depends on the values of only
a few other variables, then the structural equations become much easier
to represent. 
Normality is not typically represented using probability; in
\cite{HH11}, we represented it using a partial 
preorder;
in
\cite{Hal39,HP01b,Huber11} it is represented using a \emph{ranking
function} \cite{spohn:88}.  Both of these approaches are instances of
what has been called a \emph{plausibility measure} \cite{FrH7}.  Halpern
\citeyear{Hal25} has given conditions under which plausibility measures
can be represented using Bayesian networks; we apply these ideas here. 
This allows us to take advantage of conditional independencies to
get a compact representation of the normality order,
though it is not described probabilistically.  

We believe that even greater representational economy may often be
possible. This is because we expect that the normality order will often
be largely determined by the causal structure. For example, suppose that
the causal structure is such that if the patient suffers a head trauma,
then he would also suffer from headaches. Then we would expect any world
where trauma = 1 and headaches = 1  
to be more normal, all else being equal, than a world in which trauma = 1 and headaches = 0. 
In this way, a representation of causal structure can do ``double duty''
by representing much of the normality order as well.

An obvious question
is whether the normality order induced by the causal structure
accurately represents normality. 
This may not always be the case.
In Section~\ref{sec:Huber}, we  discuss some examples where we might
want to have a normality order that does not conform to causal
structure in this way.  
Nevertheless, we would expect that 
the normality order is
largely determined
by the equations.  Thus, we can get a more compact representation of
the normality order by just listing the \emph{exceptions} to the order
generated
by the equations.  

Interestingly, Huber \citeyear{Huber11} has suggested 
an alternative approach to representing causality and normality; 
rather than using the causal structure  to (largely) determine the
normality order, we can use the normality order to determine the
causal structure.  We discuss this possibility in more detail in
Section~\ref{sec:Huber}. 


%
  
The rest of this paper is organized as follows: In
Section~\ref{sec:extended}, we review the basic definitions needed to
understand (extended) causal models.  In Section~\ref{sec:compact}, we
discuss how compact representations of extended causal models can be
obtained; this is the technical core of the paper.  
In Section~\ref{sec:piggy-backing}, we discuss how we can (typically) get a
yet more compact representation by assuming that, by default, it is
typical for the variables to obey the structural equations.  Finally, in
Section~\ref{sec:Huber}, we discuss Huber's proposal of using the
normality order to represent causality.

\section{
Extended
Causal Models}\label{sec:extended}

Our motivation for extending causal models to incorporate a notion of 
normality is to address some difficulties facing the
Halpern-Pearl 
definition of actual cause
\cite{HP01b} and 
to extend it in various ways.
We develop the extended account in detail in
\cite{HH11}.  The Halpern-Pearl approach has
been criticized (as have all other approaches to causality!).  It is
beyond the scope of this paper to defend it.  In fact, as we shall see,
nothing in the present paper depends upon the details of the
Halpern-Pearl 
definition. Our approach to compactly representing extended causal models
can be applied to any framework that combines causal models with a 
normality ordering. 
In particular, it should be applicable to alternative accounts of actual
causation such as \cite{Hall07}.

In this section, we briefly review extended causal models.  We encourage
the reader to consult \cite{HP01b,HH11} for more details and 
motivation.
Extended causal models are based on causal models, so we start with a
review of causal models.

\subsection{Causal Models}
The description of causal models given here is taken
from \cite{Hal20}; it is a formalization of earlier work of Pearl
\citeyear{Pearl.Biometrika}, further developed in 
\cite{GallesPearl97,Hal20,pearl:2k}.

The model assumes that the world is described in terms of \emph{random
variables} and their values.  For example, if we are trying to determine
whether a forest fire was caused by lightning or an arsonist, we 
can take the world to be described by three random variables:
\begin{itemize}
\item $\FF$ for forest fire, where $\FF=1$ if there is a forest fire and
$\FF=0$ otherwise; 
\item $L$ for lightning, where $L=1$ if lightning occurred and $L=0$ otherwise;
\item $\ML$ for match dropped (by arsonist), where $\ML=1$ if the arsonist
dropped a lit match, and $\ML = 0$ otherwise.
\end{itemize}
Similarly, in an election between Mr.~B and Mr.~G with 11 voters,  we
can describe the world by 12 random variables, $V_1, \ldots, V_{11}, W$,
where $V_i = 0$ if voter 
$i$ voted for Mr.~B and $V_1 = 1$ if voter $i$ voted for Mr.~G, for $i =
1, \ldots, 11$, $W = 0$ if Mr.~B wins, and $W=1$ if Mr.~G wins.  

In these two examples, all the random variables are \emph{binary}; that
is, they take on only two values.  There is no problem allowing a random
variable to have more than two possible values.  For example, the random
variable $V_i$ could be either 0, 1, or 2, where $V_i= 2$ if $i$ does
not vote; similarly, we could take $W=2$ if the vote is tied, so neither
candidate wins.

Some random variables may have a causal influence on others. This
influence is modeled by a set of {\em structural equations}.
For example, if we want to model the fact that
if the arsonist drops a match \emph{or} lightning strikes then a fire
starts, we could use 
the random variables $\ML$, $\FF$, and $L$ as above, with the equation
$\FF = \max(L,\ML)$; that is, the value of the random variable $\FF$ is the
maximum of the values of the random variables $\ML$ and $L$.  This 
equation says, among other things, that if $\ML=0$ and $L=1$, then $\FF=1$.
The equality sign in this equation should be thought of more like an 
assignment statement in programming languages; once we set the values of
$\ML$ and $L$, then the value of $\FF$ is set to their maximum.  However,
despite the equality, if a forest fire starts some other way, that does not
force the value of either $\ML$ or $L$ to be 1.
That is, setting $\FF$ to 1 does not result in either $\ML$ or $L$ being
set to 1.

Alternatively, if we want to model the fact that a fire requires both a
lightning strike \emph{and} a dropped match (perhaps the wood is so wet
that it needs two sources of fire to get going), then the only change in the
model is that the equation for $\FF$ becomes $\FF = \min(L,\ML)$; the
value of $\FF$ is the minimum of the values of $\ML$ and $L$.  The only
way that $\FF = 1$ is if both $L=1$ and $\ML=1$.  

It is conceptually useful to split the random variables into two
sets: the {\em exogenous\/} variables, whose values are determined by 
factors outside the model, and the
{\em endogenous\/} variables, whose values are ultimately determined by
the exogenous variables.  
In the forest-fire example, the
variables $\ML$, $L$, and $\FF$ are endogenous. 
We do not want to concern ourselves with the factors
that make the arsonist drop the match or the factors that cause lightning.
Thus we do not include endogenous variables for these factors.
Instead, we introduce a single exogenous variable $U$ whose values 
take the form $(i,j)$, where $i$ and $j$ each take the value zero or one.
The value of $U$ will then
determine 
the values of $\ML$ and 
$L$.\footnote{Note that $U$ will not typically be a ``common cause''
in the usual sense. It represents a variety of different factors
which need not be correlated. Thus, we do not expect $U$ to
induce a correlation between $\ML$ and $L$.}

Formally, a \emph{causal model} $M$
is a pair $(\S,\F)$, where $\S$ is a \emph{signature}, which explicitly
lists the endogenous and exogenous variables  and characterizes
their possible values, and $\F$ defines a set of \emph{modifiable
structural equations}, relating the values of the variables.  
A signature $\S$ is a tuple $(\U,\V,\R)$, where $\U$ is a set of
exogenous variables, $\V$ is a set 
of endogenous variables, and $\R$ associates with every variable $Y \in 
\U \union \V$ a nonempty set $\R(Y)$ of possible values for 
$Y$ (that is, the set of values over which $Y$ {\em ranges}).
As suggested above, in the forest-fire example, 
we have $\U = \{U\}$, where $U$ is the
exogenous variable, $\R(U)$ consists of the four possible values of $U$ 
discussed earlier, $\V = 
\{\FF,L,\ML\}$, and $\R(\FF) = \R(L) = \R(\ML) = \{0,1\}$.  

$\F$ associates with each endogenous variable $X \in \V$ a
function denoted $F_X$ such that 
$$F_X: (\times_{U \in \U} \R(U))
\times (\times_{Y \in \V - \{X\}} \R(Y)) \rightarrow \R(X).$$
This mathematical notation just makes precise the fact that 
$F_X$ determines the value of $X$,
given the values of all the other variables in $\U \union \V$.
If there is one exogenous variable $U$ and three endogenous
variables, $X$, $Y$, and $Z$, then $F_X$ defines the values of $X$ in
terms of the values of $Y$, $Z$, and $U$.  For example, we might have 
$F_X(u,y,z) = u+y$, which is usually written as
$X\gets U+Y$.%
\footnote{Again, the fact that $X$ is assigned  $U+Y$ (i.e., the value
of $X$ is the sum of the values of $U$ and $Y$) does not imply
that $Y$ is assigned $X-U$; that is, $F_Y(U,X,Z) = X-U$ does not
necessarily hold.}  Thus, if $Y = 3$ and $U = 2$, then
$X=5$, regardless of how $Z$ is set.  

In the running forest-fire example,  where $U$ has four possible values of the form $(i,j)$, 
the $i$ value determines the value of $L$ and the $j$
value determines the value of $\ML$.  Although $F_L$ gets as arguments
the values of $U$, $\ML$, and $\FF$, in fact, it depends only
on the (first component of the) value of $U$; that is,
$F_L((i,j),m,f) = i$.  Similarly, $F_{\ML}((i,j),l,f) = j$.
In this model,
the value of $\FF$ depends only on the value of $L$ and $\ML$.
\emph{How} it depends on them depends on whether 
we are considering the conjunctive model or the disjunctive model.  


A
 \emph{
 possible world} 
 is
 an assignment of values to all the 
endogenous
 random
variables in a causal model.   
We might use the term ``small world" to describe such an assignment,
to distinguish it from a ``large world", which is an assignment of values
to all of the variables 
in a causal model, both endogenous and exogenous.

In other applications, we may well need to make use of such ``large worlds". 
In the present paper, however, we need to use 
only
small worlds.\footnote{This is because the definition of actual causation
in \cite{HP01b} involves worlds generated by performing interventions on
a causal model in a fixed \emph{context} (set of values of the exogenous
variables).  
Thus, we 
need to compare only worlds
where the
values of exogenous variables are fixed---that is, we are effectively
comparing  small worlds.}
Hence, in this paper, ``possible world" and ``world" should be
understood as referring  
to such ``small worlds". 
Intuitively, a possible world is a 
maximally specific
description of a situation 
within the language allowed by the set of endogenous variables.
Thus, a world in the forest-fire example might be one where 
$\ML = 1$, $L = 0$, and $\FF = 0$; the match is dropped, there is no
lightning, and no forest fire. 
As this example shows, a  possible world does not have to satisfy the equations
of the causal model.

A causal model $M$ is \emph{acyclic} if its equations are such that
there is no sequence of variables
$X_1$, $X_2$, \dots, $X_n$ where $X_2$ depends nontrivially on $X_1$,
\dots, $X_n$ depends non-trivially on $X_{n-1}$, and $X_1$ depends
non-trivially on $X_n$. 
In an acyclic causal model, a complete specification $\vec{U} = \vec{u}$ of 
the value(s) of the exogenous variable(s), 
called a \emph{context}, uniquely 
determines the values of all of the endogenous variables. 
In other worlds, given an acyclic causal model $M$, the choice 
of a context $\vec{u}$ suffices to determine a possible world. If the
variable $X$ takes the value $x$ in this possible world, we write $(M,
\vec{u}) \sat X = x$. 
In the sequel, we consider only acyclic causal models; these seem to be
rich enough to deal with essentially all the examples of causality that
have been considered in the literature (c.f., the discussion in
\cite{HP01b}). 

Structural equations do more than just constrain the possible values of the
variables in a causal model. They also determine what happens in
the presence of external \emph{interventions}. 
For example, we can explain what
would happen if one were to intervene to prevent the arsonist
from dropping
the match.
In the disjunctive model, there is a forest fire 
exactly if there is lightning; in the conjunctive model, there is
definitely no fire.
An ``intervention'' does not necessarily imply
human agency. The idea is rather that some independent
process overrides the existing causal structure to determine 
the value of one or more variables, regardless of the value of
its (or their) usual causes. Woodward \citeyear{Woodward03} gives 
a detailed account of such interventions. 
Setting the value of some  
endogenous
variable
$X$ to $x$ in a causal
model $M = (\S,\F)$ 
by means of an intervention
results in a new causal model denoted $M_{X
\gets x}$.  $M_{X \gets x}$ is identical to $M$, except that the
equation for $X$ in $\F$ is replaced by $X \gets x$.
Given a context $\vec{u}$, $(M_{X \gets x}, \vec{u})$ may be thought of as
the possible world that would result from intervening to set the variable $X$ to
the value $x$.
It is this ability to represent the effects of interventions on the
system that 
give a causal model its distinctively ``causal'' character.

If there are distinct values $x$, $x'$ of $X$, and $y$, $y'$ of $Y$ such
that $(M_{X \gets x}, \vec{u}) \sat Y = y$ and $(M_{X \gets {x'}},
\vec{u}) \sat Y = y'$, 
then we say that $Y$ \emph{counterfactually depends} on $X$ in 
$(M, \vec{u})$. Intuitively, this means that intervening on the value of
$X$ can make a 
difference for the value of $Y$. 

We regard a causal model as representing objective features of the world.
More precisely, given a choice of endogenous and exogenous variables,
there is a correct choice of functions to represent the causal dependence of the variables
upon one another. The correctness of a causal model can be tested, at least in principle,
by performing the relevant interventions on the values of the variables.

\subsection{Actual Causation}
One relation that has attracted considerable attention, especially in
philosophy 
and legal theory, is \emph{actual causation}. 
For example, the
claim that the arsonist's lighting his match caused the forest fire
describes a 
relation of actual causation. This claim is expressed after the fact, and it implies that
the arsonist did light his match, and that the forest fire occurred. In
addition, it 
asserts that the lit match is among the actual causes of the forest
fire. Relations of 
actual causation cannot simply be ``read off'' a causal model. Our model
of the forest  
fire tells us what would happen if the arsonist lights his match, and if lightning strikes, but
it does not tell us whether either of these events would count as a cause of the fire.
Actual causation has been of interest, in part, because it seems to be involved in
assessments of moral and legal responsibility.

The full definition of actual causation offered in \cite{HP01b} is
fairly complex, and most 
of the details do not matter to the present discussion. It suffices
to note that according to the Halpern-Pearl definition, counterfactual
dependence is \emph{sufficient} for actual causation. 
More precisely, $X = x$ is an actual cause of $Y = y$ in $(M, \vec{u})$
if (but not only if): 
(a) $(M, \vec{u}) \sat X = x, Y = y$; and (b) there exist $x' \neq x$
and $y' \neq y$ such that 
$(M_{X \gets x'}, \vec{u}) \sat Y = y'$. 
The new model $M_{X \gets x'}$, together with the context $\vec{u}$,
determines a unique value for each of the endogenous variables.
This assignment of values determines a possible world,  which is
called a \emph{witness}
to $X = x$ being an actual cause of $Y = y$.\footnote{Using the full
Halpern-Pearl definition of actual causation, 
a witness world may also incorporate the results of additional
interventions besides the intervention 
on the candidate cause. See \cite{HH11} for more details.}
Counterfactual dependence is not necessary for actual causation,
since counterfactual dependence can fail in cases of preemption and
overdetermination. But  
we can ignore these cases for now.


Halpern and Pearl 
\citeyear{HP01b}
already noted that 
a causal model does not suffice to determine causality.  There are subtle examples that can be characterized by causal models that are isomorphic, but the 
where
judgment of 
actual causation
differs.  One approach to solving these problems, 
suggested by Halpern and Pearl \citeyear{HP01b}, and developed in
different ways by 
Hall \citeyear{Hall07}, Halpern \citeyear{Hal39}, Hitchcock
\citeyear{Hitchcock07}, and Halpern and Hitchcock \citeyear{HH11}, is to
incorporate considerations about about \emph{defaults},
\emph{typicality}, and \emph{normality}. 
``Normality'' and its cognates (``normal'', ``norm'',
``abnormal'', etc.) tend to be ambiguous. They can refer to
statistical frequency, as when we say that there has been more rain than
normal for this time of year. But they can also refer to prescriptive
rules, as when we say that someone has violated a moral norm. These  
concepts obviously differ
in important ways, but in ordinary thought we often slip between the two
ideas without even realizing it. Our conjecture is that these two
different senses of ``normality'' affect causal judgments in roughly
similar ways, so we have left the word deliberately ambiguous.  
(We remark that there are other interpretations of normality as well;
see \cite{HH10} for further discussion.)

Here is a simple example to illustrate how considerations of normality can affect causal
judgments: 
\xam\label{xam:Smith}
Professor Smith and an administrative assistant take the last two
pens in the department office. There is a department rule that
administrative assistants are allowed to take the pens, while faculty
are not. Later, a problem arises from the lack of pens. 

Knobe and Fraser \citeyear{KF08} presented subjects with a version of this
vignette, and asked them to rate their agreement with \emph{either}
the statement that Professor Smith caused the problem, or that
the administrative assistant caused the problem. 
Subjects were
much more strongly inclined 
to agree
that Professor Smith caused the
problem. 

We can model this case as follows. (For simplicity, we ignore the exogenous variable(s).)
Let $PS = 1$ if Professor Smith takes a pen, $PS = 0$ if not; $AA = 1$
if the administrative assistant takes a pen, $AA = 0$ if not; and $PO =
1$ if the problem occurs, $PO = 0$ if not. Then the equation for $PO$
will be $PO = \min(PS, AA)$. It should be apparent that the dependence of
$PO$ on $PS$ and $AA$ is symmetric; in particular, $PO$ counterfactually
depends upon both variables. Nonetheless, 
judgments about the two are different.
Professor Smith violated a norm, while the administrative assistant did
not, and this difference seems to be affecting causal judgments about
the case. 

According to the theory of \cite{HH11}, potential causes are ``graded''
according to the 
normality of their witnesses.\footnote{If a potential cause has multiple witnesses, it is graded
according to its most normal witness(es).} In the pen vignette, the witness for $PS = 1$ being an
actual cause of $PO = 1$ is the world $(PS = 0, AA = 1, PO = 0)$; the witness for $AA = 1$ being
an actual cause is $(PS = 1, AA = 0, PO = 0)$. Since Professor Smith's taking a pen violates a norm, 
the former world is more normal, and $PS = 1$ receives a higher causal
grading.   
\exam

This kind of treatment can be extended to a wide range of cases. For
example, we can make 
the familiar distinction between \emph{causes} and \emph{background conditions}.
Suppose that an arsonist lit a match, oxygen was present in the air, and a fire occurred.
The fire counterfactually depends upon both the match and the oxygen,
but we tend to consider 
only the match as the cause of the fire, viewing the oxygen as a mere
background condition. 
By regarding a world with oxygen and no match as more normal than a world with a lit match and no oxygen, we can treat this case in a way that is formally analogous to the treatment of pen vignette.
Halpern and Hitchcock \citeyear{HH11} provide a number of further illustrations.

Some will worry that this account of actual causation will make
causation subjective. 
While we agree that this introduces a subjective element to actual
causation, we do not view this as a concern.
The causal model represents the 
objective core of causation. 
The patterns of causal dependence represented by the equations of a causal model are objective
features of the world. \emph{Actual causation} is a further relation that goes beyond these objective
dependence relations. It is determined \emph{in part} by objective relations of causal dependence,
but it is also determined in part by considerations of normality.

\subsection{Extended Causal Models}
Following our earlier work \cite{HH11}, we formalize normality  using \emph{extended causal models}.
We assume that there is a partial preorder $\succeq$ over worlds; 
$s \succeq s'$ means that world $s$ is at least as normal as world $s'$.
The fact that $\succeq$ is a partial preorder means that it is reflexive
(for all worlds $s$, we have $s \succeq s$, so $s$ is at least as normal
as itself) and transitive (if $s$ is at least as normal as $s'$ and $s'$
is at least as normal as $s''$, then $s$ is at least as normal as
$s''$).%
\footnote{If $\succeq$ were a partial order rather than just a partial
preorder, it would satisfy an 
additional assumption, \emph{antisymmetry}: $s \succeq s'$ and $s' \succeq s$
would have to imply $s=s'$.  This is an assumption we do \emph{not} want
to make.}  
We write $s \succ s'$ if $s \succeq s'$ and it is not the case that $s'
\succeq s$, and $s \equiv s'$ if $s \succeq s'$ and $s' \succeq s$.
Thus, $s \succ s'$ means that $s$ is strictly more normal than $s'$,
while $s \equiv s'$ means that $s$ and $s'$ are equally normal.
Note that we are not assuming that $\succeq$ is total; it is quite
possible that there are two worlds $s$ and $s'$ that are incomparable as
far as normality.  The fact that $s$ and $s'$ are incomparable does
\emph{not} mean that $s$ and $s'$ are 
equally normal.  We can interpret it as saying that the
agent is not prepared to declare either $s$ or $s'$ as more normal than
the other, and also not prepared to say that they are equally normal;
they simply cannot be compared in terms of normality.
An {\em extended causal model\/} is
a tuple $M = (\S,\F,\succeq)$, where $(\S,\F)$ is a causal model, and
$\succeq$ is a partial preorder on worlds, which can be used
to compare how normal  different worlds are. 

Partial preorders are essentially used by Kraus, Lehmann, and Magidor
\citeyear{KLM} and Shoham \citeyear{Shoham87} to model normality.
Many other approaches to modeling normality have been proposed in the
literature, including 
{\em $\epsilon$-semantics\/} \cite{Adams:75,Geffner92,Pearl90}, 
{\em possibility measures\/}
\cite{DuboisPrade:Defaults91}, and \emph{ranking functions} 
\cite{Goldszmidt92,spohn:88}.  Perhaps the most general approach uses
what are called \emph{plausibility measures} \cite{FrH7,FrH5Full}; we
return to plausibility measures below.   Some of these approaches
(specifically, 
$\epsilon$-semantics, possibilistic structures, and ranking functions)
essentially impose a total order on worlds; as we shall see, the greater
generality of partial orders provides a useful modeling tool.  That
said, almost all of what we are say in this paper applies to all these
other approaches as well.

\commentout{
According to the theory of \cite{HH11}, potential causes are ``graded''
according the 
normality of their witnesses.\footnote{If a potential cause has multiple witnesses, it is graded
according to its most normal witness(es).} In the pen vignette, the witness for $PS = 1$ being an
actual cause of $PO = 1$ is the world $(PS = 0, AA = 1, PO = 0)$; the witness for $AA = 1$ being
an actual cause is $(PS = 1, AA = 0, PO = 0)$. Since Professor Smith's taking a pen violates a norm, 
the former world is more normal, and $PS = 1$ receives a higher causal
grading.
}

\section{Compact Representations of Extended Models}\label{sec:compact}

In \cite{HH11}, a formal definition of actual causality is given; the definition is given relative to an extended causal model.   In order to determine 
actual causation
according to this definition, an agent would have to have a
representation of the model.  As we suggested in the introduction, a
na\"{\i}ve representation of a model involving $n$ binary random
variables would involve $n2^{n-1}$ values, since for each variable
$X_i$, the function $F_{X_i}$ has to give the value of $X_i$ for each of the
$2^{n-1}$ settings of the other variables.  
Even if we restrict attention to acyclic models, there may be one
variable $X$ that 
depends on all the others, so that the function $F_X$ corresponding to
$X$ has to give a value to $X$ for each of the $2^{n-1}$ settings of the
other variables.
Moreover, we must still define
a partial preorder of the $2^n$ worlds.  
Even if we restrict to total orders or use one of the other
representations of normality, since there are $2^n! \sim 2^{n
2^n}$ total orders of the worlds, this requires at least $n2^n$ bits
of information.  
Nevertheless, as we now show, in practice, it will often be possible to
represent this information in a far more compact way.

\subsection{Representing causal equations compactly: the big picture}
As we mentioned in the introduction, the key tool for getting compact
representations is the use of graphical representations.  It is
sometimes helpful to represent a causal model graphically.  
Each node in the graph corresponds to one variable in the model.
An arrow from one node, say $L$, to another, say $\FF$, indicates that
the former variable figures as a nontrivial argument in the equation 
for the latter---that is, the latter depends on the former.  Thus, we
could represent either the conjunctive or the  
disjunctive model 
of the forest-fire example
using Figure~\ref{normality-fig1}.
Note that the graph conveys only the qualitative pattern of dependence;
it does not tell us how one variable depends on others. Thus, the graph
alone does not allow us to distinguish between the disjunctive and
the conjunctive models.

\commentout{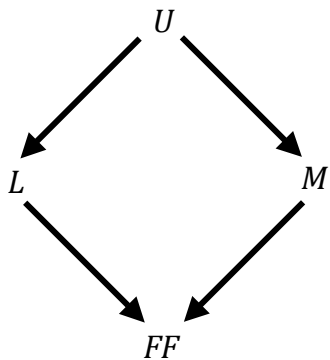
\begin{figure}[htb]
{\begin{center}
\setlength{\unitlength}{.18in}
\begin{picture}(8,9)
\put(3,0){\circle*{.2}}
\put(3,8){\circle*{.2}}
\put(0,4){\circle*{.2}}
\put(6,4){\circle*{.2}}
\put(3,8){\vector(3,-4){3}}
\put(3,8){\vector(-3,-4){3}}
\put(0,4){\vector(3,-4){3}}
\put(6,4){\vector(-3,-4){3}}
\put(3.4,-.2){$\FF$}
\put(-.7,3.8){$L$}
\put(6.15,3.8){$\ML$}
\put(3.4,7.8){$U$}
\end{picture}
\end{center}
}
\caption{A graphical representation of structural equations.}
\label{normality-fig1}
\end{figure}
}

\begin{figure}[htb]
\begin{center}
\includegraphics[width=4in]{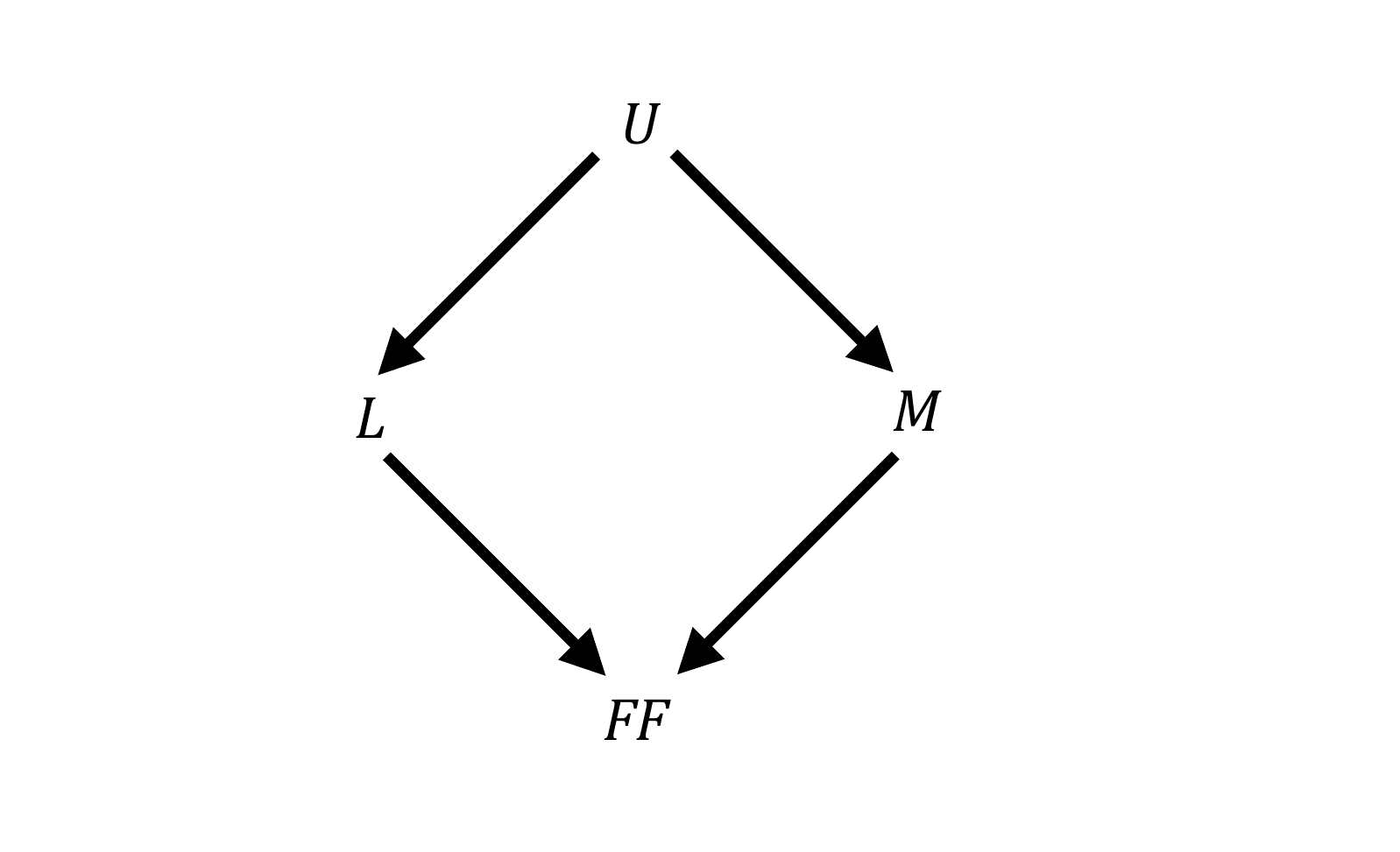}
\caption{A graphical representation of structural equations.}
\label{normality-fig1}
\end{center}
\end{figure}


The semantics (i.e., meaning) of such a graphical representation depends
on how it is being used.  In the case of causal models, it is
particularly simple.  The value of a variable in a graph depends only on
the values of its parents and is independent of the values of all other
variables once the 
values of its parents are 
given.  
Thus, in the forest-fire example, when we write the equation for $\FF$,
it depends only on the value of $L$ and $M$, and not on the value of
$U$.  (Of course, the values of $L$ and $M$ depend on the value of $U$,
so indirectly the value of $\FF$ depends on the value of $U$.) Formally,
this means that $F_{\FF}$ needs to take only two arguments (the value of
$L$ and the value of $M$), rather than 3.  More generally, if each
variable in a graph with $n$ nodes has at most $k$ parents, we can
describe the equations using $n2^k$ values.  If $k$ is small relative to
$n$ (as it is in many cases), this can be considerably less than
$n2^{n-1}$, and thus be computationally feasible. 

It is not always the case that all nodes have few parents.  Consider the voting example.   In that case, the outcome depends on how all the voters vote; that is, all of $V_1, \ldots, V_{11}$ are parents of $W$.  Thus, 
in order to describe
how $W$ 
depends upon the other variables,
we must specify
the outcome for each of the $2^{11}$  ways that the voters could vote.  
But we can do this simply without needing to list $2^{11}$ separate values:
we simply say that $W=1$ iff $V_1 + \cdots + V_{11} \ge 6$.  In the
forest-fire example, we can replace the explicit description of the
value of $\FF$ in terms of the four possible settings of $L$ and $\ML$
by just writing 
$\FF = \max(L,\ML)$ or $\FF = \min(L,\ML)$,
depending on
whether we are considering the disjunctive or conjunctive model. 
There are times when the best we can do is to write out the explicit
description of a function.  But in most cases of interest, there will be
a much more compact description.  The bottom line here is we expect to
be able to represent the structural equations compactly in most cases of
interest. 

\subsection{Representing the normality relation compactly: the big picture}


We can also make use of a graphical representation to represent the
normality order compactly. 
It is well known that \emph{Bayesian networks} can provide a compact
representation for a probability distribution.   
Suppose that we have a set $V = \{X_{1}, X_{2},\ldots, X_{n}\}$ of variables.
The worlds determined by these variables are those of the form
$(x_1, \ldots, x_n)$, where $x_i$ is a possible value of $X_i$.
If we have $n$ binary variables, and thus $2^n$ possible worlds, we need $2^n$ numbers to describe a probability distribution on these worlds.  (Actually, we need only $2^n-1$, since the sum of the numbers must be 1.)  
A \emph{quantitative Bayesian network} on a set $V = \{X_1, \ldots, X_n\}$
is an ordered pair $(G, f)$, where $G$ is a \emph{Bayesian network},
that is, a directed acyclic graph, with $n$ nodes, each labeled by a
different variable in $V$,
and $f$ associates with each variable $X_i$ in $V$ a
\emph{conditional probability table  (cpt) for $X_i$}, which describes the 
probability
that $X_i = 1$
conditional on all the possible settings of its parents 
in $G$.  
For example, if the parents of 
$X_i$ are $X_j$, $X_k$, and $X_l$, 
then we must know the probability that 
$X_i = 1$ 
conditional on each of the eight possible settings of 
$X_j$, $X_k$, and $X_l$. 
A probability measure $\Pr$ on the worlds determined by the variables in
$V$ is \emph{represented} by
$(G,f)$ if (a) $G$ satisfies the \emph{Markov 
condition}, namely, that each variable $X_i$ is independent of its
nondescendants conditional on its parents;\footnote{$Y$ is a
\emph{descendant} of $X$ if there is a directed path from $X$ to $Y$,
where we take $X$ to be a descendant of itself.  $Y$ is a nondescendant
of $X$ if it is not a descendant of $X$.  We assume that the reader is
familiar with the notion of a random variable $Y$ being independent of
another random variable $X$ conditional on a set $\mathbf{Z}$ of random
variables.  See \cite{Hal31,Pearl} for more discussion.}   
and (b) the cpt $f(X_i)$ correctly describes
the probability (according 
to $\Pr$) of $X_i$ conditional on all possible values of its parents.
If $(G,f)$ represents $\Pr$, then we can recover $\Pr$ from $(G,f)$
using quite straightforward computations (see \cite{Hal31,Pearl}).

Given a probability distribution $\Pr$, it is always possible to find a
quantitative Bayesian network $(G,f)$ that represents it.
Morever, if each node in $G$ 
has at most $k$ parents, then we need at most $(2^k-1)n$ numbers to descibe
$(G,f)$, since each cpt requires at most 
$2^k-1$ numbers.

Our representation of normality does not use probabilities; rather, it
uses a partial 
preorder
on worlds.  However, as pointed out by Halpern \citeyear{Hal25},
the ``technology'' of Bayesian networks
can be applied to mathematical structures other than
probability.  We just need a 
structure
that has a number of minimal properties and 
has an analogue of 
(conditional) independence (so that we can 
have an analogue of the Markov condition). 
Results 
of Friedman and Halpern \citeyear{FrH5Full} show that a
partial order on worlds 
gives us just such a structure.
Moreover, other representations of normality that have been considered
in the literature (e.g., ranking functions and possibility measures)
also have such a structure, so can also be represented using Bayesian
networks.  
In the language of Friedman and Halpern \citeyear{FrH7,FrH5Full}, a
sufficient 
condition for a representation of uncertainty to be represented using
Bayesian network is that it can be viewed as an \emph{algebraic
conditional plausibility measure}.  We briefly review some of the
relevant details here.

%


Plausibility measures, introduced by Friedman and Halpern
\citeyear{FrH7,FrH5Full}, are intended to be a generalization of all
standard approaches to representing uncertainty.   
The basic idea behind plausibility measures is straightforward.  A probability
measure  on a finite set $W$ of worlds maps subsets of $W$ to $[0,1]$.
A \emph{plausibility measure\/} is more general; it maps subsets of $W$  
to some arbitrary set $D$ partially ordered by $\le$.  If $\Pl$ is a
plausibility measure, $\Pl(U)$ denotes the plausibility of $U$.  If
$\Pl(U) \le \Pl(V),$ then $V$ is at least as plausible as $U$.  Because
the order is partial, it 
could be that the plausibility of two different sets is incomparable.
An agent may not be prepared to order two sets in terms of plausibility.
$D$ is assumed to contain two special elements, $\bot$ and $\top$, such
that $\bot \le d \le \top$ for all $d \in D$.  We require that
$\Pl(\emptyset) = \bot$ and $\Pl(W) = \top$.  Thus, $\bot$ and $\top$
are the analogues of 0 and 1 for probability.   We further require that
if $U \subseteq V$, then $\Pl(U) \le \Pl(V)$. 
This seems reasonable; a superset of $U$ should be at least as plausible
as $U$. 
Since Bayesian networks make such heavy use of conditioning, we need to
deal with \emph{conditional} plausibility measures (cpms), not just
plausibility measures.   A conditional plausibility measure maps pairs
of subsets of $W$ to some partially ordered set $D$.  We write $\Pl(U
\mid V)$ rather than $\Pl(U,V)$, in keeping with standard notation for
conditioning.  
We typically write just $\Pl(U)$ rather than $\Pl(U \mid W)$ (so
unconditional plausibility is identified with conditioning on the whole
space).  

In the case of a probability measure $\Pr$, 
it is standard to take $\Pr(U \mid V)$ to be undefined if $\Pr(V) = 0$.  
In general, we must make precise what the allowable second arguments of a cpm are.  For simplicity here, we assume that $\Pl(U \mid V)$ is defined as long as $V \ne \emptyset$.   For each fixed $V \ne \emptyset$, $\Pl(\cdot \mid V)$ is required to be a plausibility measure on $W$.  More generally, we require the following properties:
\begin{description}
\item[{\rm CPl1.}] $\Pl(\emptyset \mid V) = \bot$.
\item[{\rm CPl2.}] $\Pl(W \mid V) = \top$.
\item[{\rm CPl3.}] If $U \subseteq U'$, then $\Pl(U \mid V) \le \Pl(U' \mid V)$.
\item[{\rm CPl4.}] $\Pl(U \mid V) = \Pl(U \inter V \mid V)$.
\end{description}

Conditional probability satisfies additional properties; for example,
$\Pr(U\mid V') =  \Pr(U \mid V)  \times \Pr(V \mid V')$ if $U
\subseteq V \subseteq V'$ and $V \ne \emptyset$, 
and $\Pr(V_1 \union V_2 \mid  V) = \Pr(V_1 \mid V) + \Pr(V_2 \mid
V)$ if $V_1$ and $V_2$ are disjoint sets.   These properties turn out to
play a critical role in carrying out the reasoning used in Bayesian
networks.   We want plausibilistic analogues of them.  This requires us
to have plausibilistic analogues of addition and multiplication,
so that we can take $\Pl(V_1 \union V_2 \mid V_3) = \Pl(V_1\mid V_3)
\oplus \Pl(V_2 \mid V_3)$ if $V_1$ and $V_2$ are disjoint, and $\Pl(V_1
\mid V_3) = \Pl(V_1 \mid V_2) \otimes Pl(V_2 \mid V_3)$ if $V_1
\subseteq V_2 \subseteq V_3$.  We give the formal definitions in the
next section.  For now, we just note that these properties hold for 
probability if we take $\oplus$ and $\otimes$ to be $+$ and $\times$
respectively; and they hold for ranking functions if we take $\oplus$ and
$\otimes$ to be $\min$ and 
$+$.  They also hold for possibility
measures, although the situation is somewhat more complicated there
(see \cite{Hal25}).

We want to view a partial preorder on worlds as an algebraic conditional
plausibility measure.  There is a problem though.  A
plausibility measure attaches a plausibility to \emph{sets} of worlds,
not single worlds.  So, given a partial preorder $\succeq$ on worlds, we
must first find a plausibility measure $\Pl_{\succeq}$ with the property
that $\Pl_{\succeq}(w_1) \ge 
\Pl_{\succeq}(w_2)$
iff $w_1 \succeq w_2$.
We then need to show how we can extend this plausibility measure to an
algebraic conditional plausibility measure.  We do this in the next
section.  

We are mainly interested in algebraic plausibility measures 
on a set of worlds determined by random variables $X_1, \ldots, X_n$.
By results of \cite{Hal25}, such plausibility measures can be 
represented using a \emph{plausibilistic} Bayesian network $(G,f)$, where
$G$ is a Bayesian network and 
$f$ associates with each node $X_i$ in $G$ a \emph{conditional
plausibility table}; we abuse notation and use the abbreviation cpt for
a conditional plausibility table as well.
The 
cpt for $X$ specifies the plausibility of each possible value of $X$, 
conditional on all possible values of $X$'s parents.  
(Note that if $X$ is binary, it no longer
suffices to just specify the plausibility of $X=1$ conditional on $X$'s
parents, because the plausibility of $X=1$ does not necessarily
determine the plausibility of $X=0$ conditional on $X$'s parents.)
We can define what it means for a plausibilistic Bayesian network to
represent a plausiblity measure just as in the probabilistic case.  And,  
just as in the probabilistic case, a plausibility measure that takes
values in an algebraic cpm can be represented by plausibilistic Bayesian
network.  
Moreover, if each node in the network has relatively few parents, we
have a compact representation of the plausibility measure.
The bottom line here is that, once we show how to represent a partial
preorder as an algebraic plausibility measure, we can 
get a representation of the preorder using Bayesian networks that
will typically be compact.  

In addition to representing the Bayesian network, if we use a
plausibility measure, we must also represent the plausibility domain;
that is, we have to describe its elements and the ordering on them.
We expect that, typically, the domain will be relatively small, and the
ordering easy to describe.  Indeed, here the fact that we allow partial
orders makes it easier, because we allow many elements to be
incomparable.  This will become clearer in the examples in
Section~\ref{sec:examples}.

\subsection{Representing the normality relation compactly: the technical
details}\label{sec:repnorm} 

In this section, we fill in the technical details for the results
discussed in the previous section.  This section can be skipped without
loss of continuity.  We start with the formal definition of algebraic
conditional plausibility measures.

\dfn  An \emph{algebraic conditional plausibility measure} $\Pl$ on $W$
maps pairs of subsets of $W$ to a domain $D$ that is endowed with
operations $\oplus$ and $\otimes$,
defined on domains $\Dom(\oplus)$ and $\Dom(\otimes)$, respectively,
such that the following properties
hold: 
\begin{description}
\item[{\rm Alg1.}] If $V_1$ and  $V_2$ are disjoint subsets of $W$ and
$V  \ne \emptyset$, then $\Pl(V_1 \union V_2 \mid V) = \Pl(V_1 \mid V)
\oplus \Pl(V_2 \mid V)$. 
\item[{\rm Alg2.}]
If $U \subseteq V \subseteq V'$ and $V \ne \emptyset$, then 
$\Pl(U \mid V') = \Pl(U \mid V ) \otimes \Pl(V \mid V')$.

\item[{\rm Alg3.}] $\otimes$ distributes over $\oplus$; more precisely,
$a \otimes (b_1 \oplus \cdots \oplus b_n) = (a \otimes b_1) \oplus
\cdots \oplus (a \otimes b_n)$ if $(a,b_1), \ldots, (a,b_n), 
(a,b_1 \oplus \cdots \oplus b_n) \in
\Dom(\otimes)$ and $(b_1,\ldots, b_n), (a \otimes b_1,\ldots, a \otimes
b_n) \in \Dom(\oplus),$ where $\Dom(\oplus) = \{(\Pl(V_1\mid U),\ldots,
\Pl(V_n\mid U)): 
V_1, \ldots, V_n$ are pairwise disjoint and $U \ne \emptyset\}$,
and $\Dom(\otimes) = \{(\Pl(U \mid V ),\Pl(V \mid V')):
U \subseteq V \subseteq V', \, V 
\ne \emptyset\}$.  (The reason that this property is required only for 
tuples in $\Dom(\oplus)$ and $\Dom(\otimes)$ is discussed shortly.  Note
that parentheses are not required in the 
expression $b_1 \oplus \cdots \oplus b_n$ although, in general, 
$\oplus$ need not be associative. This is because it follows immediately
from Alg1 that $\oplus$ is associative and commutative on tuples in
$\Dom(\oplus)$.)  
\item[{\rm Alg4.}] If  $(a,c), \, (b,c) \in \Dom(\otimes),$
$a \otimes c \le b \otimes c$, and $c \ne \bot$, then $a \le b$.
\end{description}
\edfn
The restrictions in
Alg3 and Alg4 to tuples in $\Dom(\oplus)$ and $\Dom(\otimes)$ make these
conditions a little more awkward to state.   It may seem more natural to
consider a stronger version of, say, Alg4 that applies to all pairs in
$D \times D$. 
Roughly speaking, $\Dom(\oplus)$ and $\Dom(\otimes)$ are the only tuples
where we really care how $\oplus$ and $\otimes$ work.  We use $\oplus$
to determine the (conditional) plausibility of the union of two disjoint
sets.  Thus, we 
care about  
$\Pl(V_1 \mid V)$ and $\Pl(V_2 \mid V)$, respectively, where $V_1$ and
$V_2$ are disjoint sets, in which case we want $a \oplus b$ to be
$\Pl(V_1 \union V_2 \mid V)$.  More generally, we care about $a_1 \oplus
\cdots \oplus a_n$ only if $a_i$ has the form $\Pl(V_i \mid V)$,  where
$V_1, \ldots, V_n$ are pairwise disjoint.  $\Dom(\oplus)$ consists of
precisely 
these tuples of plausibility values.  Similarly, we care about
$a \otimes b$ only if $a$ and $b$ have the form $\Pl(U_1 \mid U_2 
)$ and $\Pl(U_2 \mid U_3)$, respectively, where $U_1 \subseteq U_2
\subseteq U_3$, 
in which case we want $a
\otimes b$ to be $\Pl(U_1 \mid U_3)$.  
$\Dom(\otimes)$ consists of
precisely these pairs $(a,b)$.
By requiring that Alg3 and Alg4
hold only for 
tuples in $\Dom(\oplus)$ and $\Dom(\otimes)$ rather than on all tuples
in $D \times D,$ some 
cpms
of interest become algebraic that would
otherwise not be.   
(See \cite{Hal25,Hal31} for examples.)
Restricting $\oplus$ and $\otimes$ to $\Dom(\oplus)$ and $\Dom(\otimes)$
will also make it easier for us to view a partial preorder as an
algebraic plausibility measure.
Since $\oplus$ and $\otimes$ are significant mainly to
the extent that Alg1 and Alg2 hold, and Alg1 and Alg2 apply to tuples in
$\Dom(\oplus)$ and $\Dom(\otimes),$ respectively, it does not
seem unreasonable that properties like Alg3
and Alg4 be required to hold only for these tuples.  

In an algebraic cpm, we can define a set $U$ to be 
\emph{plausibilistically independent of $V$ conditional on $V'$} if $V \inter
V' \ne \emptyset$ implies that $\Pl(U \mid V \inter V') = \Pl(U \mid
V')$.  The intuition here is that learning $V$ does not affect the
conditional plausibility of $U$ given $V'$.  
Note that conditional independence is, in general, asymmetric.  $U$ can
be conditionally independent of $V$ without $V$ being conditionally
independent of $U$.  
Although this may not look
like the standard definition of probabilistic conditional independence,
it is not hard to show that this definition agrees with the standard
definition (that $\Pr(U \inter V \mid V') = \Pr(U \mid V') \times \Pr(V
\mid V')$) in the special case that the plausibility measure is actually a
probability measure (see \cite{Hal25} for further discussion of this
issue).    
Of course, in this case, the definition is symmetric.


The next step is to show how to represent a partial preorder on worlds
as an algebraic plausibility measure.   
We do so using ideas of \cite{FrH5Full}. 

Suppose 
that
we have an extended causal model $M = (\S,\F,\succeq)$. 
We proceed as follows.
Define a preoder $\succeq^+$ on subsets of $W$ by taking 
$U \succeq^+ V$ if, for all $w \in V$, there exists some $w' \in U$ such that
$w' \succeq w$. 
It is easy to check that $w \succeq w'$ iff $\{w\} \succeq^+
\{w'\}$.  
Thus, we have a partial preorder on sets that extends the partial
preorder on worlds.  We might consider getting an unconditional
plausibility measure $\Pl_{\succeq}$ that extends the partial preorder
on worlds by taking the range of $\Pl_{\succeq}$ to be subsets of
$W$, and defining $\Pl_{\succeq}$ as the identity; that is, 
taking $\Pl_{\succeq}(U) = U$, and taking $U \ge V$ iff $U \succeq V$.  

This almosts works.  There is a subtle problem though.  The relation $\ge$
used in plausibility measures must be an order, not a preorder.  For an order
$\ge$ on a set $X$, if $x \ge x'$ and $x' \ge x$, we must have $x'
= x$.  Thus, for
example, if $w \succeq w'$ and $w' \succeq w$, then we want
$\Pl_{\succeq}(\{w\}) = 
\Pl_{\succeq}(\{w'\})$.  This is easily arranged.

Define $U \equiv V$ if $U \succeq^+ V$ and $V \succeq^+ U$.  
Let $[U] = \{U' \subseteq W: U' \equiv U\}$.
Now if we take $\Pl_{\succeq}(U) = [U]$, and take $[U] \ge [V]$ if 
$U' \succeq V'$ for some $U' \in [U]$ and $V' \in [V]$, then it is easy
to check that $\ge$ is well defined  
(since if $U' \succeq V'$ for some $U' \in [U]$ and $V' \in [V]$, then 
$U' \succeq V'$ for all $U' \in [U]$ and $V' \in [V]$) and is a partial
order.  

While this gives us an unconditional plausibility measure extending
$\succeq$, we are not quite there yet.  We need a conditional
plausibility measure, and a definition of $\oplus$ and $\otimes$.
Note that if $U_1, U_2
\in [U]$, then $U_1 \union U_2 \in [U]$.  Since $W$ is finite, it
follows that each set $[U]$ has a largest element, namely, the union of
the sets in $[U]$. 

Let $D$ be the domain consisting of $\bot$, $\top$, and all elements
of the form 
$d_{[U]\mid [V]}$ for all  $[U]$ and $[V]$ such that the largest element
in $[U]$ is a strict subset of the largest element in $[V]$.
We place an ordering $\ge$ on $D$ by taking $\bot < d_{[U] \mid [V]}
< \top$ and $d_{[U] \mid [V]} \le d_{[U'] \mid [V']}$ if $[V] = [V']$ and
$U' \succeq^+ U$.  
%
We view
$D$ as the range of an algebraic plausibility measure, defined by
taking $$\Pl_{\succeq}(U \mid V) = 
\left\{
\begin{array}{ll}
\bot &\mbox{if $U \inter V   = \emptyset$}\\ 
\top &\mbox{if $U \inter V = V$}\\ 
d_{[U \inter V] \mid [V]} &\mbox{otherwise.} 
\end{array}
\right.
$$
We can define $\oplus$ and $\otimes$ on $D$ so that
Alg1 and Alg2 hold.
This is easy to do, in large part because we only need to define
$\oplus$ and $\otimes$ on $\Dom(\oplus)$ and $\Dom(\otimes)$, where the
definitions are immediate because of the need to satisfy Alg1 and Alg2.
It is easy to see that these conditions and the definition of $\Pl$
guarantee that Pl1--4 hold.  With a little more work, it can be shown that 
these conditions imply Alg3 and Alg4 as well. (Here the fact that Alg3
and Alg4 are restricted to $\Dom(\oplus)$ and $\Dom(\otimes)$ turns out
to be critical; it is also important that 
$U \equiv V$ implies that $U \equiv
U \union V$.)  

This construction gives us an algebraic cpm, so the results of 
\cite{Hal25} apply.  
%
Specifically, we can represent $\succeq$ using
a Bayesian network  
$(G, f)$. 
The structure of $G$ is determined by the independencies exhibited
by the normality order on worlds. There is no guarantee that
$G$ will be the same as the graph that represents the causal structure.
In many cases, however, there will be substantial overlap between 
the Bayesian network representation 
of the normality order and the causal model.
As we show in Section~\ref{sec:piggy-backing} below, 
when this occurs, even greater economy is possible.

\subsection{Using a compact representation to determine a normality order}
\label{sec:examples}

The discussion above shows that if we start with an algebraic conditional
plausibility measure determined by a normality
order on worlds, then we can represent it using a Bayesian network.
Moreover, this representation will often be compact.  But what we really
want is more like the converse.  Suppose that we are given a
quantitative Bayesian network.  Can we use that to determine a normality
order on worlds?   The reason that we are particularly interested in
this question is that, in many cases of interest, 
it is quite natural to characterize a situation using a
quantitative Bayesian network.

Suppose, for example, that a lawyer is arguing that a defendant should be
convicted of arson. The lawyer will attempt to establish a claim of actual
causation: that the defendant's action of lighting a match was an
actual cause of the forest fire. To do this, she will need to convince
the jury that a certain extended causal model is correct, and that
certain initial conditions 
obtained (for example, that the defendant did indeed light a match). 
To justify a causal model, she will need to defend the equations. This might
involve convincing the jury that the defendant's match was the sort of
thing that could cause a forest fire (the wood was dry), and that there
would have been no fire in the absence of some triggering event, such as
a lightning strike or an  act of arson.  

The lawyer will also have to defend a normality ordering.
To do this, she might argue that a lightning strike could not have
been reasonably foreseen; that lighting a fire in the forest at that time
of year was in violation of a statute; and that it was to be expected that 
a forest fire would result from such an act. The key idea here is that
it will usually be easier to justify a normality ordering in a 
\emph{piecemeal} fashion. Instead of arguing for a particular 
normality ordering on entire worlds, she argues that individual 
variables typically take certain values in certain 
situations.\footnote{Here and subsequently
we make use of an artificial terminological convention introduced 
in \cite{HH11}. We use ``typical" and its cognates when talking about 
individual variables. For example, we say that it is atypical for lightning
to strike. We reserve ``normal" and its cognates for comparisons of 
entire worlds. Formally, however, both are represented by plausibility 
values.} In doing this, she is defending a particular cpt for each variable.
What we show in this section is that having a cpt for each variable
leads in a natural way to a particular choice of normality ordering on
worlds.  

\commentout{
In the probabilistic setting, a Bayesian network $(G,f)$ does indeed
determine a unique probability distribution.  The same is true more
generally in the plausibilistic setting, if we make the appropriate
assumptions.  More precisely, suppose that we start with a   
\emph{BN-compatible} domain $D$, that is, a plausibility domain $D$ 
(i.e., a partially preordered set with $\top$ and $\bot$ such that $\bot
\le d \le \top$ for all $d \in D$), binary operators $\oplus$ and
$\otimes$ on $D$ satisfying a number of natural conditions.
Then given a Bayesian network
$(G,f)$ where each value in the cpt that $f$
associates with a node $X$ in $G$ is in $D$, there is a unique
plausibility measure with values in $D$ determined by $(G,f)$.
Moreover, this plausibility measure can be computed in exactly the same
way as the probablity measure (by using $\otimes$ and $\oplus$ instead
of $\times$ and $+$).  (See \cite{Hal25,Hal31} for details.)

But this is still not quite what we want.  It would require much more
than just a quantitative Bayesian network.  We would need to define a
BN-compatible domain, which would mean defining $\oplus$ and $\otimes$,
and checking a number of properties.  It turns out that since all we want
at the end of the day is a normality order on worlds, we can take a
much simpler approach, which requires only the Bayesian network.  We now
sketch how this can be done.
}

Recall that the Bayesian network $G$ is labeled by variables $X_1,
\ldots, X_n$; we want to define a normality order on worlds of the
form $(x_1, \ldots, x_n)$, where $x_i$ is a possible value of the random
variable $X_i$.  We will associate with each such world a plausibility
value of the form $a_1 \otimes \cdots \otimes a_n$, where $a_i$ is a
value in the cpt for $X_i$.  For example, if $n=3$, and, according to 
$G$, $X_1$ and $X_2$ are independent and $X_3$ depends on both $X_1$ and
$X_2$, then a world $(1,0,1)$ would be assigned a plausibility of 
$a_1 \otimes a_2 \otimes a_3$, where $a_1$ is the unconditional
plausibility of $X_1
= 1$ according to the cpt for $X_1$, $a_2$ is the unconditional
plausibility of $X_2=0$ according to the cpt for $X_2$, and $a_3$ is the
plausibility of $X_3 =1$ conditional on $X_1 = 1 \inter X_2 =0$, given
by the cpt for $X_3$.  We do not need to actually define 
the operation $\otimes$ here; we
just 
leave
$a_1 \otimes a_2 \otimes a_3$ as
an uninterpreted expression.
However, if we have some constraints on the
relative order of elements in the cpt (as we do in our examples, and
typically will in practice), then lift this to an order on
expressions of the form $a_1 \otimes a_2 \otimes a_3$ by taking
$a_1 \otimes a_2 \otimes a_3 \le a_1' \otimes a_2' \otimes
a_3'$ 
if and only if,
for all $a_i$, there exists some $a_j'$ such that $a_i \le
a_j'$.  
The ``only if'' builds in a
minimality assumption: two elements 
$a_1 \otimes a_2 \otimes a_3$ and $a_1' \otimes a_2' \otimes
a_3'$
are incomparable 
unless they are forced to be comparable 
by the ordering relations among the $a_i$'s and the $a_j'$'s.
One advantage of using a partial preorder, rather than a total preorder,
is that we can do this.

These assumptions determine a unique partial preorder on elements of the
form $a_1 \otimes \cdots \otimes a_n$.  This gives us a partial preorder
on worlds.  (We can then use the construction in Section
\ref{sec:repnorm} to then obtain an algebraic plausibility measure 
that in fact represented by $(G,f)$,
but this is no longer necessary, since
all we care about is the normality order on worlds.)  

While the formal foundations of our approach involve some complexities,
the application of these ideas to specific cases is often quite intuitive. 
The following two examples 
show how this construction might work in our running example.


\xam\label{xam:Fire} Consider the forest-fire example again.  Here we
can take the worlds to have the form 
$(i,j,k)$
where 
$i$, $j$, and $k$ are the values of 
$\ML$, $L$, and $\FF$,
respectively.  
We can represent the independencies in the forest-fire example using the
network in Figure~\ref{normality-fig1}
(with $U$ removed).
Thus, $L$ and $\ML$ are
independent, and $\FF$ depends on both of them.

For definiteness, consider the disjunctive case, where either a
lightning strike or an arsonist's match suffices for fire.  
It would be natural to say that lightning strikes and  
arson attacks are atypical, and that a forest fire typically occurs if
either of these events occurs.  
Suppose that we use $d_L^+$ to represent the plausibility of $L=0$ (lightning
not occuring) and $d_L^-$ to represent the plausibiltiy of $L=1$;
similarly, we use $d_{\ML}^+$ to represent the plausibility of $\ML=0$ and
$d_{\ML}^-$ to represent the plausibility of $\ML=1$.
Now the question is what we should take the conditional plausibility of
$\FF=0$ and $\FF=1$ to be given each of the four possible settings of $L$
and $\ML$.  For simplicity, we take all the four values compatible with the
equations to be equally plausible, and have plausibility $d_{\FF}^+$,
and the values incompatiable with the equations to all be equally
plausible and have plausibility $d_{\FF}^-$.  
This gives us the following cpts:
\begin{equation}\label{eq:cpt}
\begin{array}{ll}
\Pl(L = 0) = d_L^+\, > d_{L}^-\ = \Pl(L = 1)\\ 
\Pl(\ML = 0) = d_{M}^+ > d_M^- = \Pl(\ML = 1)\\ 
\Pl(\FF = 0 \mid L = 0 \land \ML = 0) = d_{\FF}^+ > d_{\FF}^- =  \Pl(\FF = 1 \mid L
= 0 \land \ML = 0)\\ 
\Pl(\FF = 1 \mid L = 1 \land \ML = 0) = d_{\FF}^+ >  d_{\FF}^- = \Pl(\FF = 0 \mid L = 1
\land \ML = 0)\\ 
\Pl(\FF = 1 \mid L = 0 \land \ML = 1) = d_{\FF}^+ > d_{\FF}^- = \Pl(\FF = 0 \mid L =
0 \land \ML = 1)\\
\Pl(\FF = 1 \mid L = 1 \land \ML = 1) = d_{\FF}^+ > d_{\FF}^- =  \Pl(\FF = 0 \mid L =
1 \land \ML = 1).
\end{array}
\end{equation} 

\begin{figure}[htb]
\begin{center}
\includegraphics[width=4.5in]{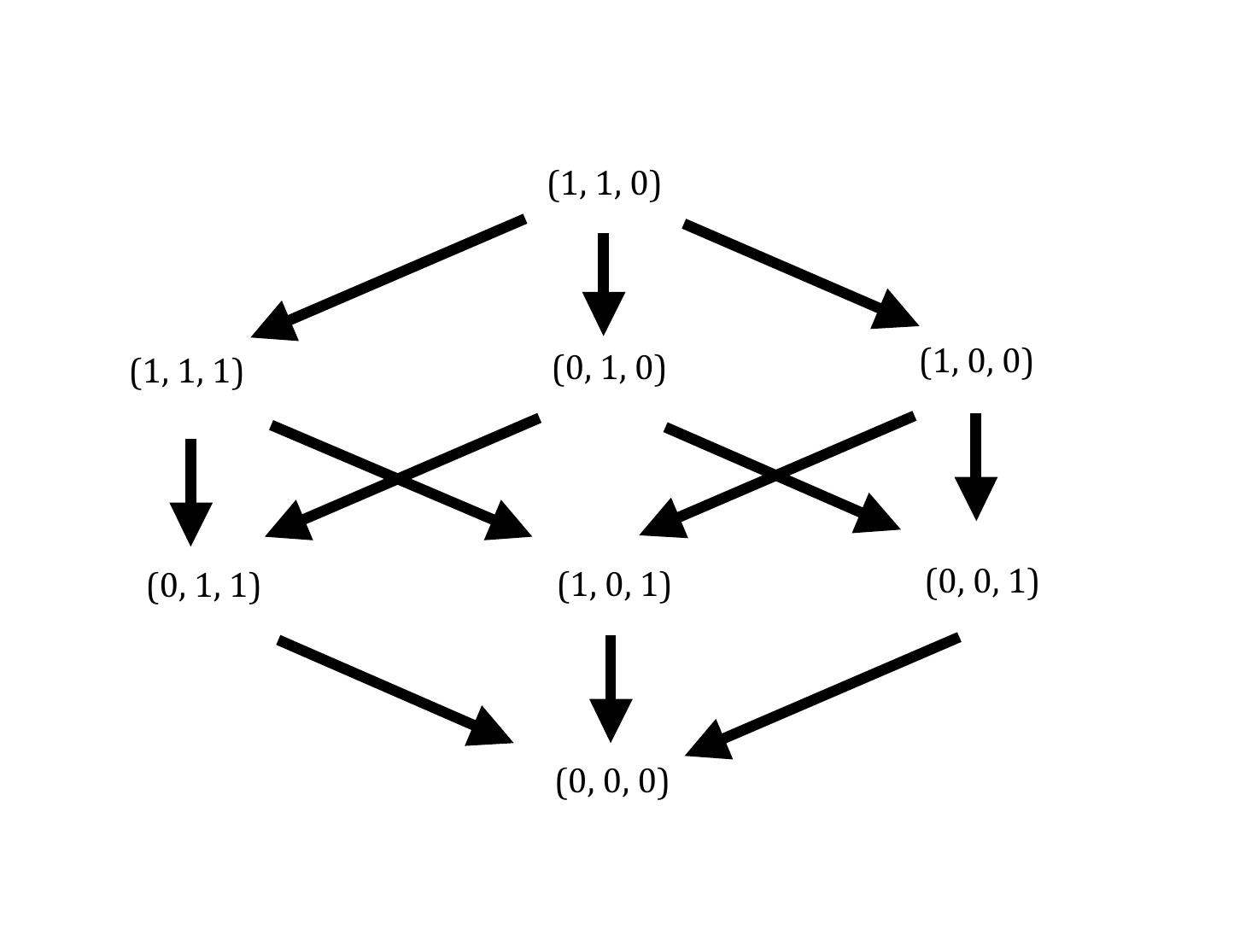}
\caption{A normality order on worlds.}
\label{network1}
\end{center}
\end{figure}

Suppose that we further assume that $d_L^+$, $d_M^+$, and $d_{\FF}^+$
are all incomparable, as are $d_L^-$, $d_M^-$, and $d_{\FF}^-$.  Thus,
for example, we cannot compare the degree of typicality of no lightning
with that of no arson attacks, or the degree of atypicality of lightning
with that of an arson attack.  
Using the construction above gives us the the ordering on 
worlds 
given in Figure~\ref{network1},
where an arrow from $w$ to $w'$ indicates that $w' \succ w$.

In this normality order, (0, 1, 1) is more normal than (1, 1, 1) and (0, 1,
0), but incomparable with (1, 0, 0) and (0, 0, 1). 
That is because, according to our construction, $(0,1,1)$, $(1,1,1)$, 
$(0,1,0)$, $(1,0,0)$, and $(0,0,1)$ have plausibility 
$d_L^+ \otimes d_M^- \otimes d_{\FF}^+$,
$d_L^- \otimes d_M^- \otimes d_{\FF}^+$,
$d_L^+ \otimes d_M^- \otimes d_{\FF}^+$,
$d_L^- \otimes d_M^+ \otimes d_{\FF}^-$, and
$d_L^+ \otimes d_M^+ \otimes d_{\FF}^+$, respectively.
The fact that $d_L^+ \otimes d_M^- \otimes d_{\FF}^+ \ge
d_L^- \otimes d_M^- \otimes d_{\FF}^+$ follows since $d_L^+ \ge d_L^-$.
The fact that we have $>$, not just $\ge$, follows from the fact that we
do \emph{not} have $d_L^- \otimes d_M^- \otimes d_{\FF}^+ \ge
d_L^+ \otimes d_M^- \otimes d_{\FF}^+$, since this does not follow from
our condition from comparability.  The other comparisons follow from
similar arguments.
\exam

\commentout{
We can capture this order relation by constructing cpts
for the three variables $L$, $\ML$, and $\FF$.
For example, we could consider a simple domain of plausibilities that includes 
elements $\bot$, $\top$, $d_L^+$, $d_L^-$, $d_{\FF}^+$, $d_{\FF}^-$, 
$d_{M}^+$, and $d_M^-$, with $\bot < d_X^- < d_X^+ < \top$ for $X \in
\{L,M,\FF\}$, but otherwise all these elements are incomparable, and
assume that the cpts are given as follows:
}

\begin{figure}[htb]
\begin{center}
\includegraphics[width=4.5in]{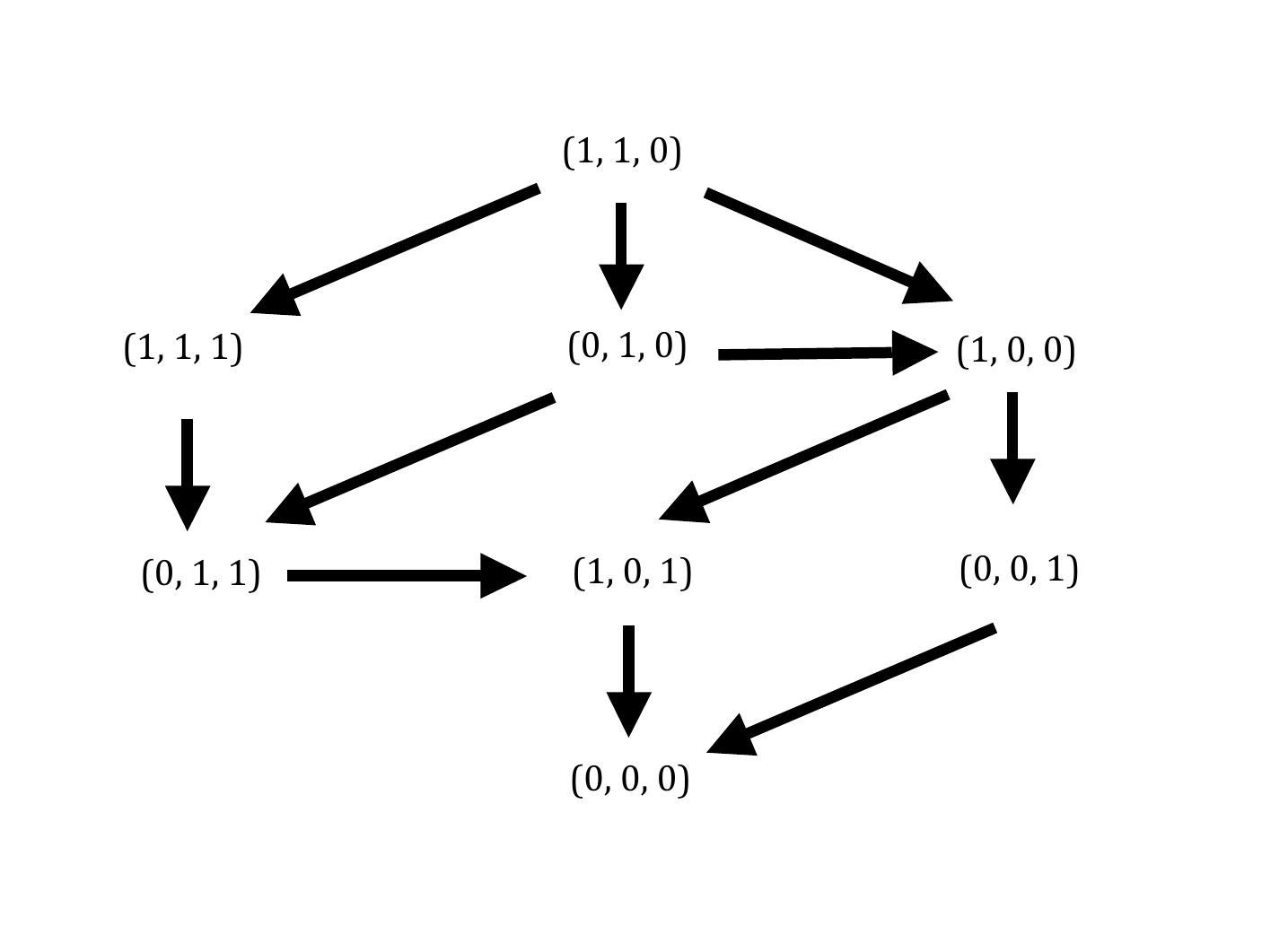}
\caption{A different normality order on worlds.}
\label{network}
\end{center}
\end{figure}

\xam\label{xam:Comparability}
The order on worlds induced by the Bayesian network in the previous
example treats the lightning and the 
arsonist's actions as incomparable. For example, the world $(1, 0, 1)$,
where lightning strikes, the arsonist doesn't, and there is a fire, is
incomparable with the world where lightning doesn't strike, the arsonist
lights his match, and the fire occurs. But this is not the only
possibility. Suppose that we judge that it would be more atypical for
the arsonist to light a fire than for lightning to strike, and also more
typical for the arsonist not to light a fire than for lightning not to
strike. (Unlike the case of probability, the latter does not follow from
the former.)  
Recall that this order might reflect the fact that arson is illegal
and immoral, rather than the 
frequency of occurrence of arson as opposed to lightning.
While (\ref{eq:cpt}) still describes the conditional plausibility
tables, we now have $d_{L}^+ > d_M^+$ and $d_M^- >d_L^-$.
This gives us the order on worlds described in Figure~\ref{network}.

Now, for example, the world $(0, 1, 1)$ is strictly more normal than the
world $(1, 0, 1)$; 
again, the former has plausibility $d_L^+ \otimes d_M^- \otimes
d_{\FF}^+$, while the latter has 
plausibility $d_L^- \otimes d_M^+ \otimes d_{\FF}^+$.  But since $d_L^+
> d_L^-$ and $d_L^+ > d_M^+$, by assumption, it follows that 
$d_L^+ \otimes d_M^- \otimes d_{\FF}^+ > d_L^- \otimes d_M^+ \otimes
d_{\FF}^+$.  
\exam

\commentout{
\subsection{Examples}\label{sec:examples}
%
}

\section{Piggy-Backing on the Causal Model}\label{sec:piggy-backing}

If the normality order is represented by a Bayesian network $(G,
f)$, there is no guarantee that the graph $G$ will duplicate the graph
corresponding to the causal model. Nonetheless, in many cases it will be
reasonable to expect that there will be substantial agreement between
the two graphs.  
When this happens, it will be possible to make parts of the causal model
do ``double duty'': representing both causal structure, and the structure
of the normality order.  
In Examples~\ref{xam:Fire} and~\ref{xam:Comparability}, the graph
describing the causal structure is the same as the graph in the
Bayesian network representation of the normality order. This 
reflects the idea that a fire typically occurs when causes of a fire are
present.  

But we can say more than this.   
Consider the conditional plausibility table for the variable $\FF$ 
from (\ref{eq:cpt}).  We can summarize it as follows:
$$\Pl(\FF = \mathit{ff} \mid L = l \land \ML = m) = 
\left\{
\begin{array}{ll}
d_{\FF^+} &\mbox{if $\mathit{ff} = \max(l,m)$}\\ 
d_{\FF^-} &\mbox{otherwise.} 
\end{array}
\right.
$$
%
Recall that $\FF = \max(L,\ML)$ is the structural equation for $\FF$ in
the causal model. So the conditional plausibility table says, in effect,
that it is typical for $\FF$ to obey the structural equations, and
atypical to violate it.  

Variables typically obey the
structural equations.  Thus, it is often far more efficient to assume
this holds by default, and explicitly enumerate cases where this is not
so, rather than writing out all the equations.  Specifically, we propose
the following default rule.
\begin{quote}
\textbf{Default Rule 1} (\emph{Normal Causality}): 
Let $X$ be a variable in a causal model
with no exogenous parents, and let
$\mathbf{PA}(X)$ be the 
vector of parents of $X$. Let the structural equation for $X$ be $X =
f_X(\mathbf{PA}(X))$. 
Then, unless explicitly given otherwise, there are two plausibility
values $d_X^+$ and $d_X^-$ with $d_X^+ > 
d_X^-$ such that 
$$\Pl(X = x \mid \mathbf{PA}(X) = \mathbf{p_X}) = 
\left\{
\begin{array}{ll}
d_{X^+} &\mbox{if $x = f_X(\mathbf{p_X})$}\\ 
d_{X^-} &\mbox{otherwise.} 
\end{array}
\right.
$$
\end{quote}
Default Rule 1 tells us that it is typical for variables to satisfy the
equations, unless we 
explicitly stipulate otherwise.
In Examples~\ref{xam:Fire} and ~\ref{xam:Comparability}, $\FF$ satisfies
Default Rule 1.
Moreover, it says that, by default, all 
%
values of variables
that satisfy the
equations are equally typical, while all those that do not satisfy the
equations are equally atypical.  Of course, we could allow some
deviations from the equations to be more atypical than others; this
would be a violation of the default rule.  
As the name suggests,
the default rule is to be assumed, unless
explicitly stated otherwise. 
The hope is that there will
be relatively few violations, so there is still substantial
representational economy in assuming the rule.
That is, the hope is that, once a causal model is given,
the normality order can be represented efficiently by providing the
conditional plausibility tables for only those variables that violate
the default rule,
or whose plausibility values are not determined by the default rule
(because they have exogenous parents).\footnote{It may be possible
to formulate more complex versions of Default Rule 1 that accommodate
exogenous parents, and allow for more than two default values.
We leave these extensions for another occasion.}

The \emph{Normal Causality} rule, by itself, does not tell us how the
plausibility values in the cpt for one variable 
compare to the plausibility values in the cpt for another
variable. We therefore supplement our first default rule with a second: 

\begin{quote}
\textbf{Default Rule 2} (\emph{Minimality}): If $d_x$ and $d_y$ are
plausibility values in the conditional plausibility table for distinct
variables $X$ and $Y$ and no information is given explicitly regarding
the relative 
orders of $d_x$ and $d_y$, then $d_x$ and $d_y$ are incomparable. 
\end{quote}
Again, this default rule is assumed to hold only if there is no explicit
stipulation to the contrary. 
Default Rule 2 tells us that the normality ordering among possible worlds
should not 
include any comparisons that do not follow from the equations 
(via Default Rule 1) together with the information that is explicitly
given.\footnote{
Roughly speaking, in the context of probability,
a distribution that maximizes entropy 
subject to some constraints is one that 
is (very roughly) the one that makes things ``as equal as possible''
subject to the constraints.  If there are no
constraints, it reduces to
the classic principle of 
indifference, which tells 
us to assign equal probability to different possibilities in the absence
of any reason to think 
some are more probable.  In the
context of plausibility, where only weak order is assigned, 
it is possible to push this idea a step further by making the
possibilities incomparable.}  
In Example~\ref{xam:Fire}, all three variables satisfy \emph{Minimality}. 
In Example~\ref{xam:Comparability}, $\FF$ satisifies \emph{Minimality}
with respect to 
the other two variables, but the variables $L$ and $\ML$ do not satisfy it with respect to 
one another (since their values are stipulated to be comparable). 

With these two default rules, 
we can represent the extended causal model 
in Example~\ref{xam:Fire} 
succinctly as follows:
$$\begin{array}{l}
\FF = \max(L,\ML)\\
\Pl(L = 0) > \Pl(L = 1)\\
\Pl(\ML = 0) > \Pl(\ML = 1).
\end{array}$$
The rest of the structure of the normality order follows from the default rules. 

In Example~\ref{xam:Comparability}, we can represent the extended causal
model as follows: 
$$\begin{array}{l}
\FF = \max(L,\ML)\\ 
\Pl(\ML = 0) > \Pl(L = 0) > \Pl(L = 1) > \Pl(\ML = 1).
\end{array}$$
Again, the rest of the structure follows from the default rules. In each
case, the normality order  
among the eight possible worlds can be represented with the addition of
just a few plausibility values 
to the causal model. Thus, moving from a causal model to an extended
causal model need not  
impose enormous cognitive demands.

Exceptions to the default rules can come in many forms. 
There could be values of the variables for which violations of the
equations are more typical than agreements with the equations. 
As we we suggested after Default Rule 1, there
could be multiple values of typicality, 
rather than just two for each variable.%
\footnote{Note that there are many structural equations for a
variable $X$.  Indeed, if $X$ has $k$ parents, there are $2^k$
equations, one for each possible setting of the values of the parents of
$X$.  An equation like $\FF = \max(L,\ML)$ packages up the four
equations into one compact equation.  Default Rule 1 assumes that all
agreements with these $2^k$ equations get a plausibility of $d_X^+$, and
all violations get a plausibility of $d_X^-$.  But we could certainly
view some violations as less typical than others.}
Or the
conditional plausibility values of one variable could be comparable with
those of another variable.  
These default rules are useful to the extent that there are relatively few
violations of them.
For some settings, other default
rules may also be useful; the two rules we have presented are certainly
not the only possible useful defaults.



\section{Nothing but Normality?}\label{sec:Huber}

In a recent paper, Huber \citeyear{Huber11} claimed 
that it is unnecessary to employ distinct modalities for
normality and causal structure, and that it is preferable to encompass
both in a unified normality structure. Huber's framework employs a
family of ranking functions to represent normality. Huber shows that if
the ranking functions satisfy a condition that he calls ``respect for the
equations'', one can use the ranking functions as a ``similarity metric''
on possible worlds, and give a semantics for counterfactuals in the
spirit of  
Stalnaker68 \citeyear{Stalnaker68}
or Lewis \citeyear{Lewis73}.
In this way, all the information about counterfactuals is already
contained in the ranking functions; it is unnecessary to give the
structural equations as a distinct element of the model. Moreover, this
semantics provides truth values for propositions in a richer language
than that of  
Galles and Pearl \citeyear{GallesPearl98},
Halpern \citeyear{Hal20}, 
or 
Briggs \citeyear{Briggs12}.
In particular, it yields truth values for embedded counterfactuals, where the antecedent of a counterfactual conditional includes a counterfactual. 

Huber's requirement that the ranking functions respect the equations is
similar in spirit to 
the \emph{Normal Causality} default rule,
in that it requires 
worlds that violate more equations to receive higher rank. 
(Higher rank corresponds to lower plausibility.) It is a bit more
complicated than this, since it also gives priority to worlds where
violations occur ``later'', as measured by number of steps in a directed
path. This is supposed to ensure that the closest possible world to $w$
in which some variable $X$ takes a value $x$ different from the one it
takes in $w$, is one where $X$ takes the value $x$ due to a ``miracle''
that occurs as late as possible.  

We do not go into all of the details of Huber's result here. It is easy to
see that if there is a plausibility measure that satisfies 
\emph{Normal Causality}
or something similar, then there is a natural sense in which the
structural equations are encoded in that plausibility measure. Huber's
result is one specific way of making of this idea precise.  

Huber's result is both interesting  and technically impressive. Nonetheless, we prefer to retain causal structure and a normality order as distinct modalities. We have this preference for 
several
reasons. 

First, while Huber's result provides a kind of conceptual unification,
it is not at all clear that it provides a more compact representation.
Indeed, as we have argued, it is often possible to provide a representation
of the causal structure plus the normality ordering that is very compact. 
\commentout{it does not provide for a 
more compact representation of an extended causal model. Huber's framework
requires a separate ranking function for each context. In the
forest-fire example, this means 
that one would have to specify four separate ranking functions (one for
each value of the exogenous variable $U$, or equivalently, one for each
combination of values of $L$ and $\ML$).  
It is hard to see how this could be represented as compactly as the representations suggested in the previous section.
}

Second,
we think that the normality ordering and the causal model are conceptually
representing very different things.
The causal
structure, as represented in the equations of a causal model, is an
objective feature of a system. For example, the accuracy of a causal
model can be evaluated by performing appropriate observations and
interventions on the system. By contrast, normality can be affected by
social rules, moral norms, and the like. The normality order may reflect
features of the way in which an agent reasons about a system, but it is
not something that can be confirmed experimentally. 
We believe that actual causation
involves both of these components;
it
is partly objective and partly
value-laden. 
Our framework keeps these two distinct components separate
and makes explicit the different roles they play in judgments of actual 
causation.
Finally,
there are examples where normality and causal structure do and should come apart. 
Huber briefly discusses this point at the very end of his paper. He
concludes that we should not rely on mere intuitions about normality in
cases such as these, but should instead put weight on the conceptual
economy and unification that results in his framework. As we now show,
however, there are some examples where the cleaving of normality and
causal structure is justified not only by intuition,  
but also by the demands of a theory of actual causation.

Recall Example~\ref{xam:Smith}, in which Professor Smith and the
administrative 
assistant took the two remaining pens. We had three endogenous variables:
$PS$, representing whether or not Professor Smith takes a pen; $AA$, representing
whether or not the administrative assistant takes a pen; and $PO$ representing whether
or not a problem occurs. To capture the judgments of the subject in the experiment,
we want it to be atypical for Professor Smith to take the pen ($PO = 1$).
Let us now suppose that we add a further variable to our model: $CP = 1$ 
if the department chair institutes a policy forbidding faculty members from taking
pens; $CP = 0$ if she does not institute such a policy. (We could add additional
possible values corresponding to alternative policies, such as forbidding everyone
from taking pens, but this is not essential to the present point.) How does the new
variable $CP$ relate to $PS$? On the one hand, it seems that $CP$ influences
which value of $PS$ is typical. When $CP = 1$, 
Professor Smith's taking a pen violates a norm. 
But it is also apparent that the chair's policy had no effect on
Professor Smith; 
he took a pen despite the policy (let us assume that he willfully ignored the 
policy). Thus, we want our extended model to say \emph{both}
that Professor Smith would take the pen if the chair implements the policy,
\emph{and} that this violates a norm. We cannot do this if the same
ordering is used for both normality and the structural equations. 

\section{Conclusion}
The goals of this paper are relatively modest.  We highlight a
problem that we believe has not been considered in the causality
literature, and propose a solution to it.  We believe that any
reasonable approach to causality must pass a minimal ``psychological
feasibility''
test; the models must be representable compactly.  We
have shown that this can be done in practice with the Halpern-Pearl
model, and with other 
approaches
that involve structural equations and
possibly also a normality ordering.  We believe that such compactness
considerations should be taken into account in any attempt to model human
reasoning; far too often in the philosophical literature, it has not
been considered.

\bibliographystyle{harvard}
\bibliography{z,joe,refs}

\end{document}

%% file: defn.tex
\newtheorem{THEOREM}{Theorem}[section]
\newenvironment{theorem}{\begin{THEOREM} \hspace{-.85em} {\bf :} }%
                        {\end{THEOREM}}
\newtheorem{LEMMA}[THEOREM]{Lemma}
\newenvironment{lemma}{\begin{LEMMA} \hspace{-.85em} {\bf :} }%
                      {\end{LEMMA}}
\newtheorem{COROLLARY}[THEOREM]{Corollary}
\newenvironment{corollary}{\begin{COROLLARY} \hspace{-.85em} {\bf :} }%
                          {\end{COROLLARY}}
\newtheorem{PROPOSITION}[THEOREM]{Proposition}
\newenvironment{proposition}{\begin{PROPOSITION} \hspace{-.85em} {\bf :} }%
                            {\end{PROPOSITION}}
\newtheorem{DEFINITION}[THEOREM]{Definition}
\newenvironment{definition}{\begin{DEFINITION} \hspace{-.85em} {\bf :} \rm}%
                            {\end{DEFINITION}}
\newtheorem{CLAIM}[THEOREM]{Claim}
\newenvironment{claim}{\begin{CLAIM} \hspace{-.85em} {\bf :} \rm}%
                            {\end{CLAIM}}
\newtheorem{EXAMPLE}[THEOREM]{Example}
\newenvironment{example}{\begin{EXAMPLE} \hspace{-.85em} {\bf :} \rm}%
                            {\end{EXAMPLE}}
\newtheorem{REMARK}[THEOREM]{Remark}
\newenvironment{remark}{\begin{REMARK} \hspace{-.85em} {\bf :} \rm}%
                            {\end{REMARK}}

\newcommand{\thm}{\begin{theorem}}
\newcommand{\lem}{\begin{lemma}}
\newcommand{\pro}{\begin{proposition}}
\newcommand{\dfn}{\begin{definition}}
\newcommand{\rem}{\begin{remark}}
\newcommand{\xam}{\begin{example}}
\newcommand{\cor}{\begin{corollary}}

\newcommand{\ethm}{\end{theorem}}
\newcommand{\elem}{\end{lemma}}
\newcommand{\epro}{\end{proposition}}
\newcommand{\edfn}{\bbox\end{definition}}
\newcommand{\erem}{\bbox\end{remark}}
\newcommand{\exam}{\bbox\end{example}}
\newcommand{\ecor}{\end{corollary}}

\newcommand{\beqn}{\begin{equation}}
\newcommand{\eeqn}{\end{equation}}

\newcommand{\bbox}{\vrule height7pt width4pt depth1pt}

\newcommand{\clm}{\begin{claim}}
\newcommand{\eclm}{\end{claim}}

\newcommand{\sat}{\models}

\newcommand{\union}{\cup}
\newcommand{\inter}{\cap}

\newcommand{\FF}{{\bf F}}

\renewcommand{\phi}{\varphi}

\newcommand{\F}{{\cal F}}

\newcommand{\R}{{\cal R}}

\newcommand{\U}{{\cal U}}
\newcommand{\V}{{\cal V}}

\newcommand{\ol}{\setlength{\itemsep}{0pt}\begin{enumerate}}
\newcommand{\eol}{\end{enumerate}\setlength{\itemsep}{-\parsep}}
\newcommand{\ul}{\setlength{\itemsep}{0pt}\begin{itemize}}
\newcommand{\dl}{\setlength{\itemsep}{0pt}\begin{description}}
\newcommand{\edl}{\end{description}\setlength{\itemsep}{-\parsep}}
\newcommand{\eul}{\end{itemize}\setlength{\itemsep}{-\parsep}}

\newcommand{\commentout}[1]{}

\newcommand{\bi}{\begin{itemize}}
\newcommand{\ei}{\end{itemize}}
\newcommand{\be}{\begin{enumerate}}
\newcommand{\ee}{\end{enumerate}}